\documentclass[conference]{IEEEtran}
\IEEEoverridecommandlockouts
\usepackage{cite}
\usepackage{amsmath,amssymb,amsfonts}
\usepackage{algorithmic}
\usepackage{graphicx}
\usepackage{textcomp}
\usepackage{xcolor}
\def\BibTeX{{\rm B\kern-.05em{\sc i\kern-.025em b}\kern-.08em
    T\kern-.1667em\lower.7ex\hbox{E}\kern-.125emX}}
\usepackage{latexsym}
\usepackage{subcaption}
\usepackage{booktabs}
\usepackage{multirow}
\usepackage{diagbox}
\usepackage{colortbl}
\usepackage{bbding}
\definecolor{mygray}{gray}{.9}

\begin{document}

\title{Few-shot Object Detection with Refined Contrastive Learning}

\author{
\IEEEauthorblockN{Zeyu Shangguan, Lian Huai$^{\ast}$ \thanks{*Corresponding author}, Tong Liu, Xingqun Jiang}
\IEEEauthorblockA{\textit{AIoT CTO} \\
\textit{BOE Technology Group Co., Ltd.}\\
Beijing, China \\
\{shangguanzeyu, huailian, liutongcto, jiangxingqun\}@boe.com.cn}
% \and
% \IEEEauthorblockN{}
% \IEEEauthorblockA{\textit{AIoT CTO} \\
% \textit{BOE Technology Group Co., Ltd.}\\
% Beijing, China \\
% huailian@boe.com.cn}
% \and
% \IEEEauthorblockN{3\textsuperscript{rd} Tong Liu}
% \IEEEauthorblockA{\textit{AIoT CTO} \\
% \textit{BOE Technology Group Co., Ltd.}\\
% Beijing, China \\
% liutongcto@boe.com.cn}
% \and
% \IEEEauthorblockN{4\textsuperscript{th} Xingqun Jiang}
% \IEEEauthorblockA{\textit{AIoT CTO} \\
% \textit{BOE Technology Group Co., Ltd.}\\
% Beijing, China \\
% jiangxingqun@boe.com.cn}
}

\maketitle

\begin{abstract}
Due to the scarcity of sampling data in reality, few-shot object detection (FSOD) has drawn more and more attention because of its ability to quickly train new detection concepts with less data. However, there are still failure identifications due to the difficulty in distinguishing confusable classes. We also notice that the high standard deviation of average precision reveals the inconsistent detection performance. To this end, we propose a novel FSOD method with Refined Contrastive Learning (FSRC). A pre-determination component is introduced to find out the Resemblance Group from novel classes which contains confusable classes. Afterwards, Refined Contrastive Learning (RCL) is pointedly performed on this group of classes in order to increase the inter-class distances among them. In the meantime, the detection results distribute more uniformly which further improve the performance. Experimental results based on PASCAL VOC and COCO datasets demonstrate our proposed method outperforms the current state-of-the-art research.
\end{abstract}


\section{Introduction}
\label{sec:intro}

With the emergence of deep learning and convolutional neural network, object detection in computer vision and image processing fields has been significantly improved. In late studies, few-shot object detection (FSOD) is mainly trained in the two-stage fine-tuning based approach \cite{Wang20} using Faster R-CNN \cite{Ren15}. In the first stage, the baseline model is pre-trained with dataset containing a large amount of data such as PASCAL VOC \cite{Everingham15} and COCO \cite{cocodataset}. In the second stage, the classification and regression layers are used to fine-tune the system with limited data to further improve the performance. 

\begin{figure}[ht]
	\centering
	\begin{subfigure}{0.3\linewidth}
		\centering
		% \fbox{\rule{0pt}{2in} \rule{.9\linewidth}{0pt}}
		\includegraphics[width=25mm]{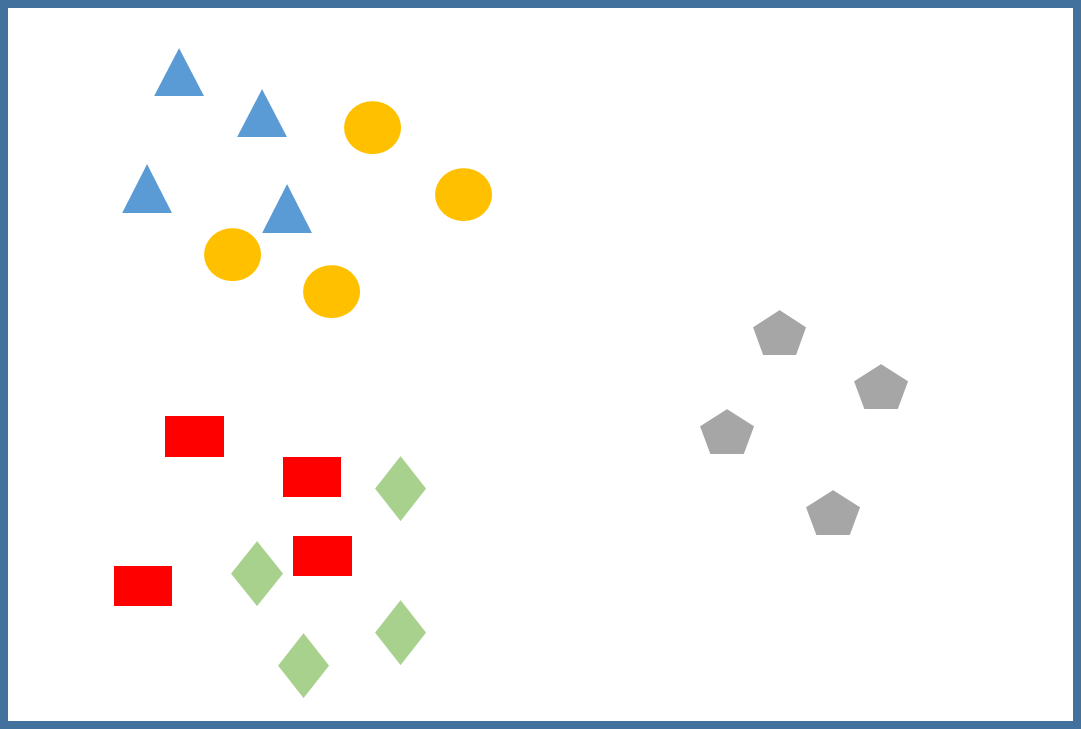}
		\caption{}
		\label{fig:intro-a}
	\end{subfigure}
	\begin{subfigure}{0.3\linewidth}
		\centering
		% \fbox{\rule{0pt}{2in} \rule{.9\linewidth}{0pt}}
		\includegraphics[width=25mm]{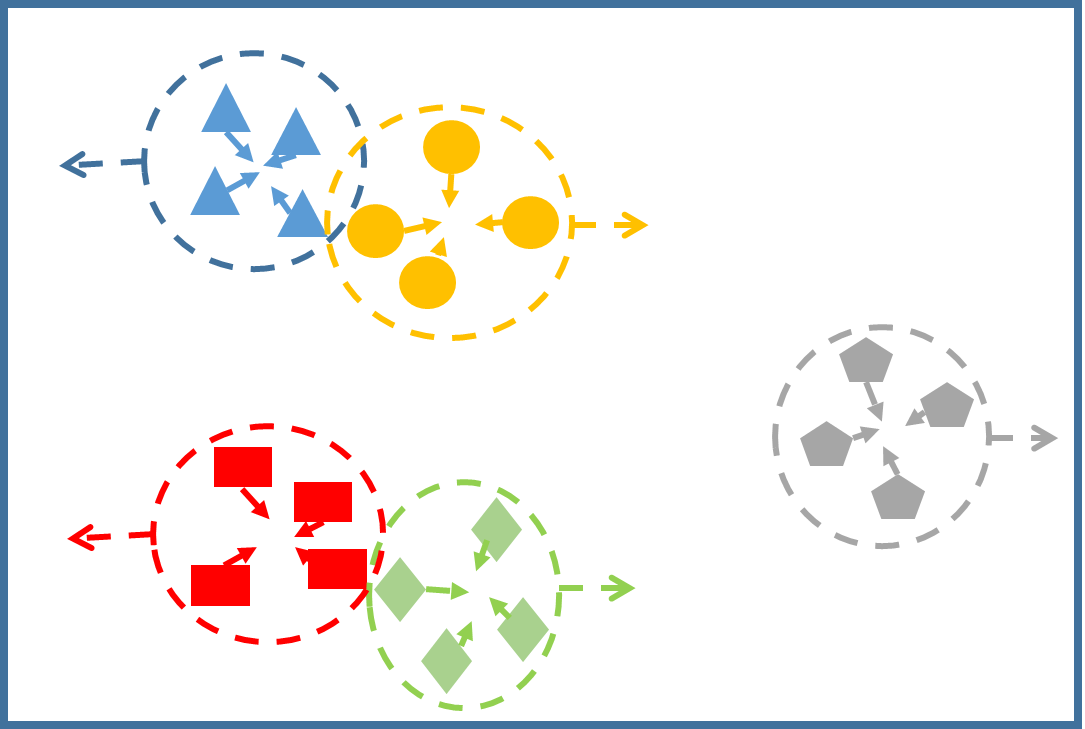}
		\caption{}
		\label{fig:intro-b}
	\end{subfigure}
	\begin{subfigure}{0.3\linewidth}
		\centering
		% \fbox{\rule{0pt}{2in} \rule{.9\linewidth}{0pt}}
		\includegraphics[width=25mm]{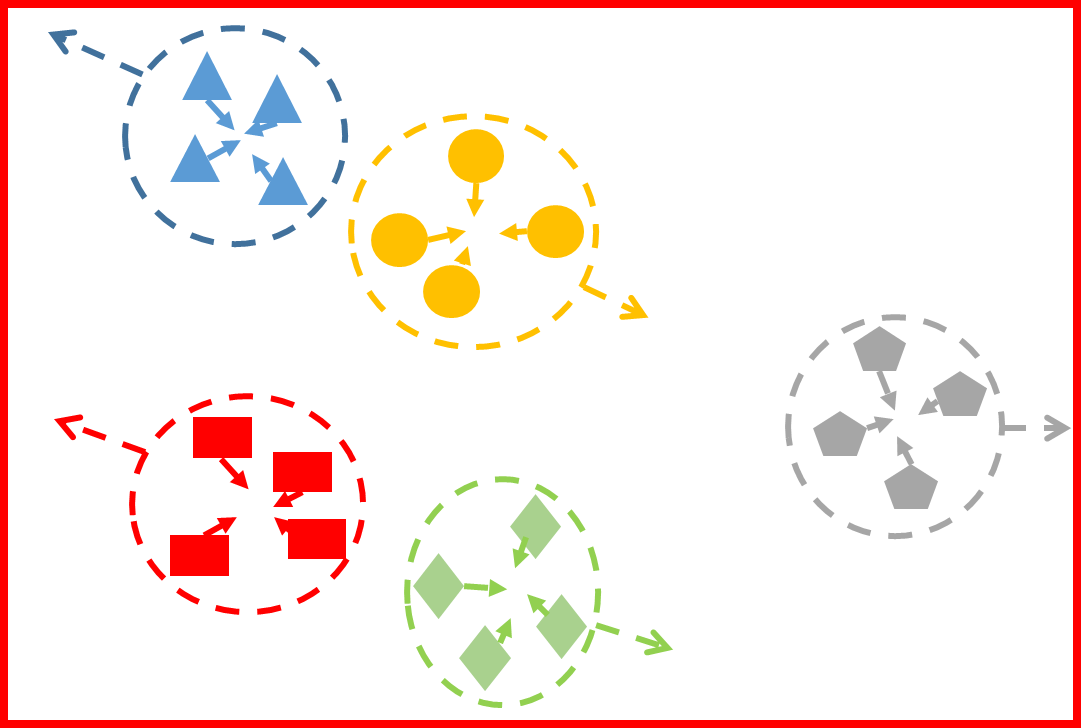}
		\caption{}
		\label{fig:intro-c}
	\end{subfigure}

	\begin{subfigure}{\linewidth}
		\centering
		% \fbox{\rule{0pt}{2in} \rule{.9\linewidth}{0pt}}
		\includegraphics[width=77mm]{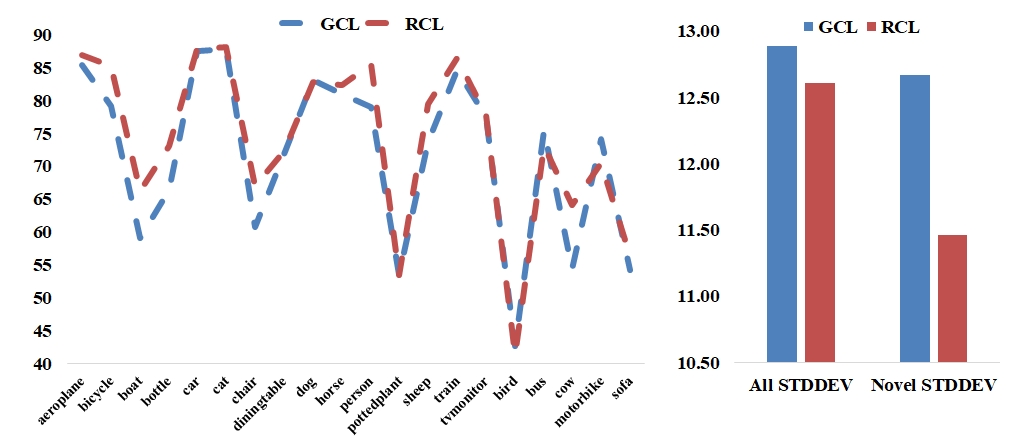}
		\caption{}
		\label{fig:intro-d}
	\end{subfigure}
	\caption{Part (a) to (c) are conceptualization of non-contrastive learning (CL), general CL (GCL) and our proposed refined CL (RCL) respectively, in which our RCL presents a more balanced inter-class distance than the other two. In part (d),the dotted lines on the left side shows our RCL's AP50 outperforms GCL on most of the classes. The right side display the standard deviation of AP50 for novel classes and all classes, and our RCL have apparently lower standard deviation compared to GCL.}
	\label{fig:intro}
\end{figure}

\begin{figure*}[ht]
	\centering
	\begin{subfigure}{0.26\linewidth}
		\centering
		% \fbox{\rule{0pt}{2in} \rule{.9\linewidth}{0pt}}
		\includegraphics[height=28mm]{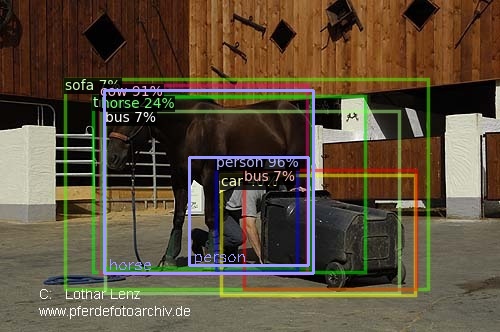}
		\caption{Horse vs. *cow}
		\label{fig:example_01}
	\end{subfigure}
	\begin{subfigure}{0.26\linewidth}
		\centering
		% \fbox{\rule{0pt}{2in} \rule{.9\linewidth}{0pt}}
		\includegraphics[height=28mm]{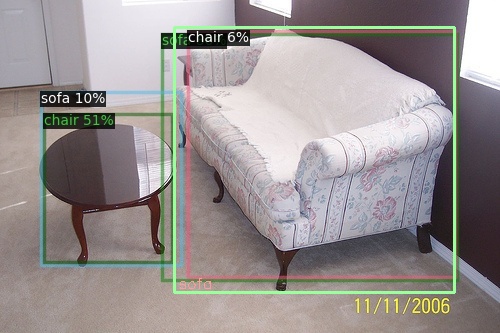}
		\caption{Chair vs. *sofa}
		\label{fig:example_02}
	\end{subfigure}
	\begin{subfigure}{0.26\linewidth}
		\centering
		% \fbox{\rule{0pt}{2in} \rule{.9\linewidth}{0pt}}
		\includegraphics[height=28mm]{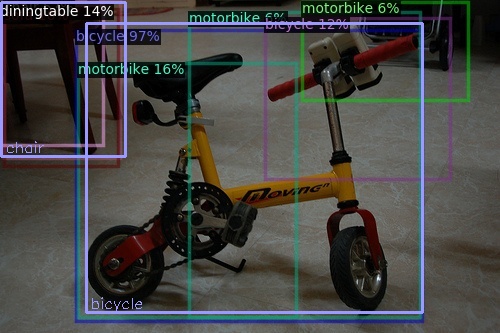}
		\caption{Bicycle vs. *motorbike}
		\label{fig:example_03}
	\end{subfigure}
	\begin{subfigure}{0.16\linewidth}
		\centering
		% \fbox{\rule{0pt}{2in} \rule{.9\linewidth}{0pt}}
		\includegraphics[height=28mm]{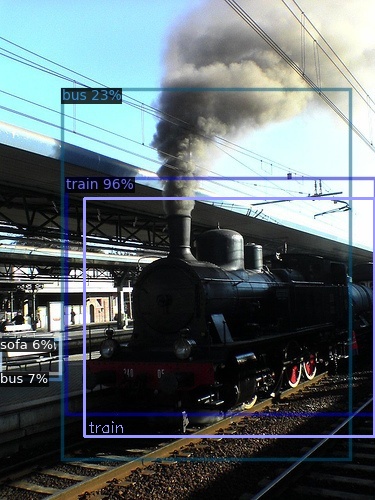}
		\caption{Train vs. *bus}
		\label{fig:example_04}
	\end{subfigure}
	\caption{Example of resemblance pairs from PASCAL VOC. Ground truth boxes are in solid shape with labels at the lower left corner. Predicted boxes are translucent with labels at the upper left corner. Sign * indicates novel classes.}
	\label{fig:example}
\end{figure*}

Moreover, in many cases, the high variance of the limited training data leads to unreliable detection results on existing benchmarks. Reference \cite{Wang20} introduces a revised evaluation protocol by sampling multiple groups of training data sets to attain more stable results. More complicated feature aggregation and meta-learning on a more balanced data set has significantly improved the performance. However for novel classes, the degradation of mean average precision (mAP50) that results from misclassification still exist. To mitigate this, \cite{Sun21} promotes an instance level intra-class compactness and inter-class variance via contrastive proposal encoding loss.

Albeit the mainly degradation comes from misclassifying are improved in \cite{Sun21} by considering confusable classes in novel instances, we notice that the contrastive learning acts uniformly on all the classes so that the efforts are diluted by those non-confusable classes. Consequently there is still a bipolar phenomenon of average precision (AP50) scores for different classes. Therefore, inspired by the supervised contrastive learning \cite{Khosla20}, we propose a few-shot object detection based on refined contrastive learning (FSRC) to further improve the mAP50 of the model, and ease the bipolar phenomenon of AP50. To the best of our knowledge, we are the first to report the standard deviation of AP50 for both base and novel data when assessing the reliability of predictions for different classes.

As shown in Fig.~\ref{fig:intro}, our proposed approach focuses on enlarging the inter-class distances in resemblance group rather than averagely magnifying the inter-class distances as general contrastive learning (GCL) does. Our proposed refined contrastive learning (RCL) automatically finds out the resemblance group (i.e. a subset of all target categories that contains confusable classes, defined in Section.\ref{sec: Refined contrastive learning (RCL)}) by searching resemblance pairs (as shown in Fig.~\ref{fig:example}) and focuses on this group at the later stages of fine-tune training. The effect is obvious in terms of reducing mAP50 and smoothing the variances among different classes' AP50. Efforts are not made on classes which are already easy to be classified in our system, but more training overhead are put on confusable classes in resemblance group. 

\begin{figure*}[ht]
	\hfill
	\begin{subfigure}{\linewidth}
		\centering
		% \fbox{\rule{0pt}{2in} \rule{.9\linewidth}{0pt}}
		\includegraphics[width=145mm]{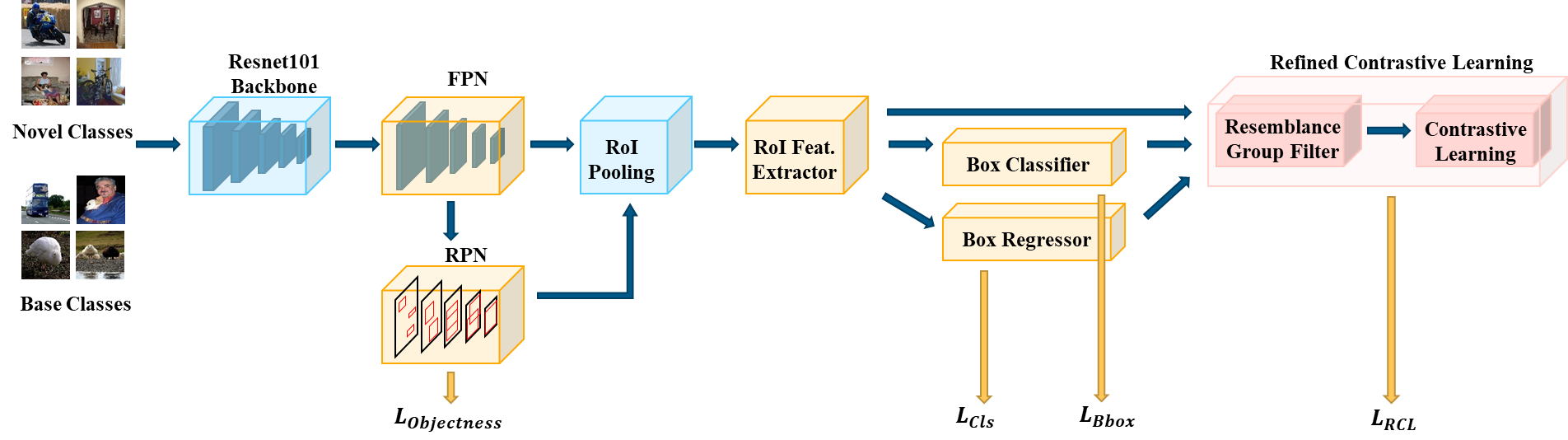}
		\caption{}
		\label{fig:flowchart-a}
	\end{subfigure}
	\hfill
	\begin{subfigure}{\linewidth}
		\centering    
		% \fbox{\rule{0pt}{2in} \rule{.9\linewidth}{0pt}}
		\includegraphics[width=145mm]{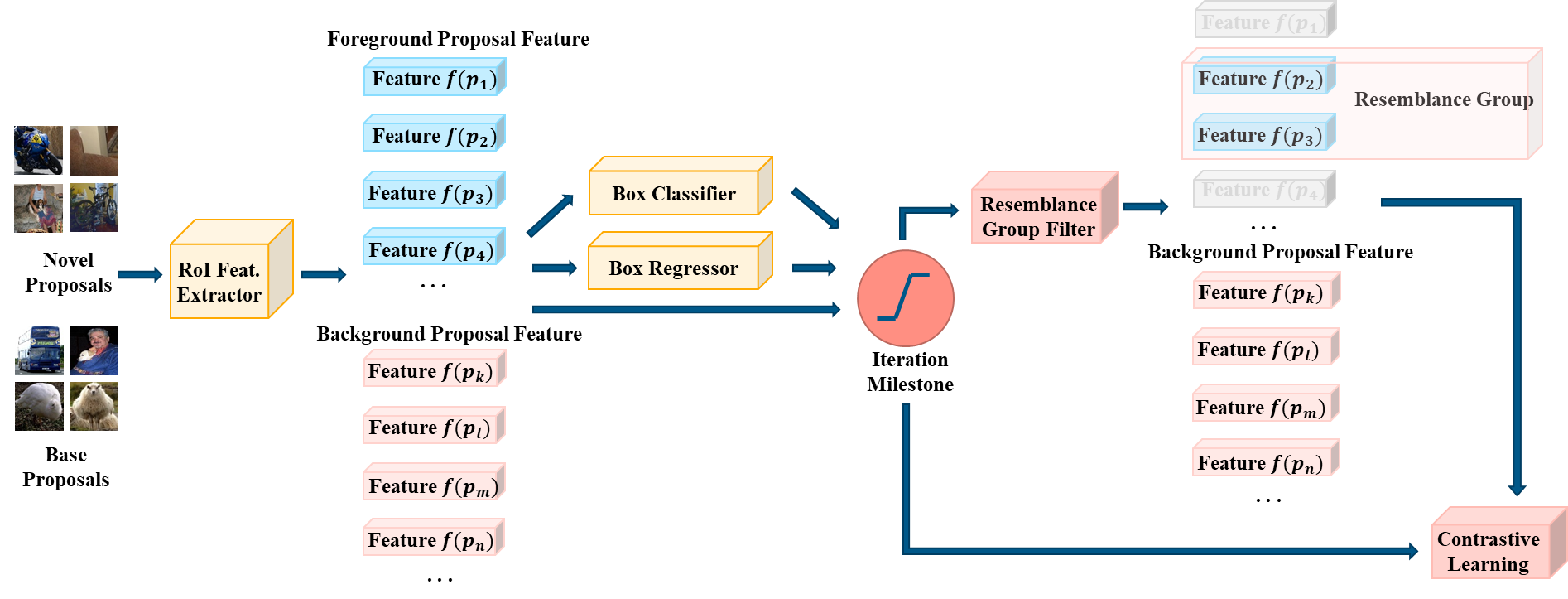}
		\caption{}
		\label{fig:flowchart-b}
	\end{subfigure}
	\caption{Our proposed FSRC. Part (a) is the overview of FSRC pipeline. A special refined contrastive learning (RCL) is added after the RoI pooling layer. Part (b) illustrates the details of our RCL head. Before the iteration percentage hit the milestone, we apply general contrastive learning. Afterwards, the resemblance group is filtered out and sent to RCL.}
	\label{fig:flowchart}
\end{figure*}

\section{Related work}
\label{sec:related work}

\subsection{Few-shot learning object detection}
Recently, FSOD algorithms are mostly divided into two branches: meta-learning and fine-tuning. A meta-learner is trained to help transferring knowledge learned from base classes and quickly adapt to new tasks. Reference \cite{Wang20} uses complex feature aggregation and meta-training with a balanced data set. Reference \cite{Sun21} improves upon \cite{Wang20} by integrating contrastive learning in FSOD to mitigate the degradation of mAP50. Reference \cite{Fan21} gives a conclusion of quantitative analysis that unfreezing the RPN in fine-tuning is beneficial to FSOD since the RPN is class-agnostic. Reference \cite{Li21c} decouples the classification and regression and claim that the classification is translation-invariant while regression is translation-covariant.

%-------------------------------------------------------------------------
\subsection{Contrastive learning}

As a self-supervised learning technology, contrastive learning learns the feature representation by measuring the feature similarities between samples\cite{Tian20,Gao21}. Reference \cite{Wu18} achieve competitive results in instance level classification. Reference \cite{Zhu22} use contrastive learning in long-tail visual recognition task. For few-shot learning, \cite{Luo21,Yang22} applies contrastive learning in few-shot classification. For the first time, \cite{Sun21} introduces contrastive learning in few-shot object detection. Reference \cite{Huang22} applies contrastive learning in multi-model few-shot object detection.

\section{Proposed approach}
\label{sec:proposed approach}
%-------------------------------------------------------------------------
\subsection{Observation}

By checking on the mean average precision (mAP50) of all novel classes, we notice that the training results under general fine-tuning scenario have a fairly uneven performance: some classes' average precision (AP50) could reach 0.8, while others remain as low as 0.2. However, such a polarization phenomenon seems not due to unbalanced sampling and obeys the data distribution since images are selected randomly and evenly from each classes. FSCE \cite{Sun21} tries to apply general supervised contrastive learning to increase the inter-class distances, however, such a jagged AP50 is still severe. Therefore, we intend to overcome this explicit polarization of the performance of the novel classes. We reconsider the relationship among all classes, including both base class ($C_{base}$) and novel class ($C_{novel}$). We assume that there exist classes which share similar features. To this end, we visualize the predicting results and prove the existence of this group of classes, which we call it resemblance group ($G_{R}$) in our paper. Some typical resemblance pairs in the $G_{R}$ are shown in Fig.~\ref{fig:example}.

\subsection{Baseline}

We follow the baseline network denoted by \cite{Wang20}. The overview structure of our proposed FSRC network in Fig.~\ref{fig:flowchart-a} is comprised of three major components. The first component is the backbone that consists of feature pyramid network (FPN) and region proposal network (RPN) used for extracting features from input images and obtaining region proposals. The second part is the region of interest (RoI) feature extraction module, which regrinds proposals in box classifier and box regressor to get detection results. The third part is our major innovation which contributes to the improvements. Fig.~\ref{fig:flowchart-b} illustrates the details of RCL. Resemblance group ($G_{R}$) is filled up with resemblance pairs found from foreground batch proposals. Proposals in $G_{R}$ are then sent to contrastive learning along with the background proposals.

%-------------------------------------------------------------------------
\subsection{Refined contrastive learning (RCL)}
\label{sec: Refined contrastive learning (RCL)}

In general supervised contrastive learning, if two classes naturally have large inter-class distance, contrastive learning between them would not make an eminent improvement precision wise. Meanwhile, this process might distract contrastive learning from working on classes with small inter-class distance, which are supposed to be the major target of contrastive learning. Our RCL is capable of redressing this unfocused effect. The $G_{R}$ with small inter-class distance is picked out by automatic statistics. After $G_{R}$ is generated, only those proposed boxes whose predicted classification or ground truth classification belong to $G_{R}$ will be sent to contrastive learning. This is also clearly shown in Fig.~\ref{fig:flowchart-b}. The major steps are described as follows.

%-------------------------------------------------------------------------
{\bf Find the resemblance group.} Proposals from FPN are fed into the RoI head to carry out further classification and sent to box regressor to localize instances. To find out the $G_{R}$, the predicted proposal boxes from RoI head are examined by the following steps:

{\bf (\romannumeral1) $\mathbf{IoU}$ threshold.} In order to identify the most frequently misclassified classes, we set an appropriate $IoU$ threshold ($Th_{IoU}^{G_{R}}$) to pick out the foreground proposals that have the highest IoU overlap with their ground truth boxes. Only the proposals fulfill the following two requirements are reserved into the $G_{R}$: (1) The proposal's $IoU$ with the ground truth box ($IoU_{p}$) is larger than the $IoU$ threshold ($IoU_{p}\!>\!Th_{IoU}^{G_{R}}$). (2) The predicted classification result $C^{i}_{p}$ ( where $i$ indicates $i$th proposal) is different from its ground truth $C^{j}_{gt}$ ( where $j$ represents $j$th ground truth box), i.e. $C^{i}_{p}\!\neq\!C^{j}_{gt}$. Then we record the two classes as a resemblance pair ($P_{R}\!=\!\left\langle C^{i}_{p}, C^{j}_{gt}\right\rangle$). Otherwise, the proposal is ignored and will not be sent to contrastive learning. For instance, in Fig.~\ref{fig:example_03}, boxes of motorbike that satisfy the $Th_{IoU}^{G_{R}}$ with the ground truth box of bicycle, then they are recorded as a resemblance pair; while box of dining table can not satisfy the $Th_{IoU}^{G_{R}}$ requirement with the ground truth box of bicycle and thus they will not be recorded as resemblance pair.

\begin{figure}
	\centering
	\includegraphics[width=85mm]{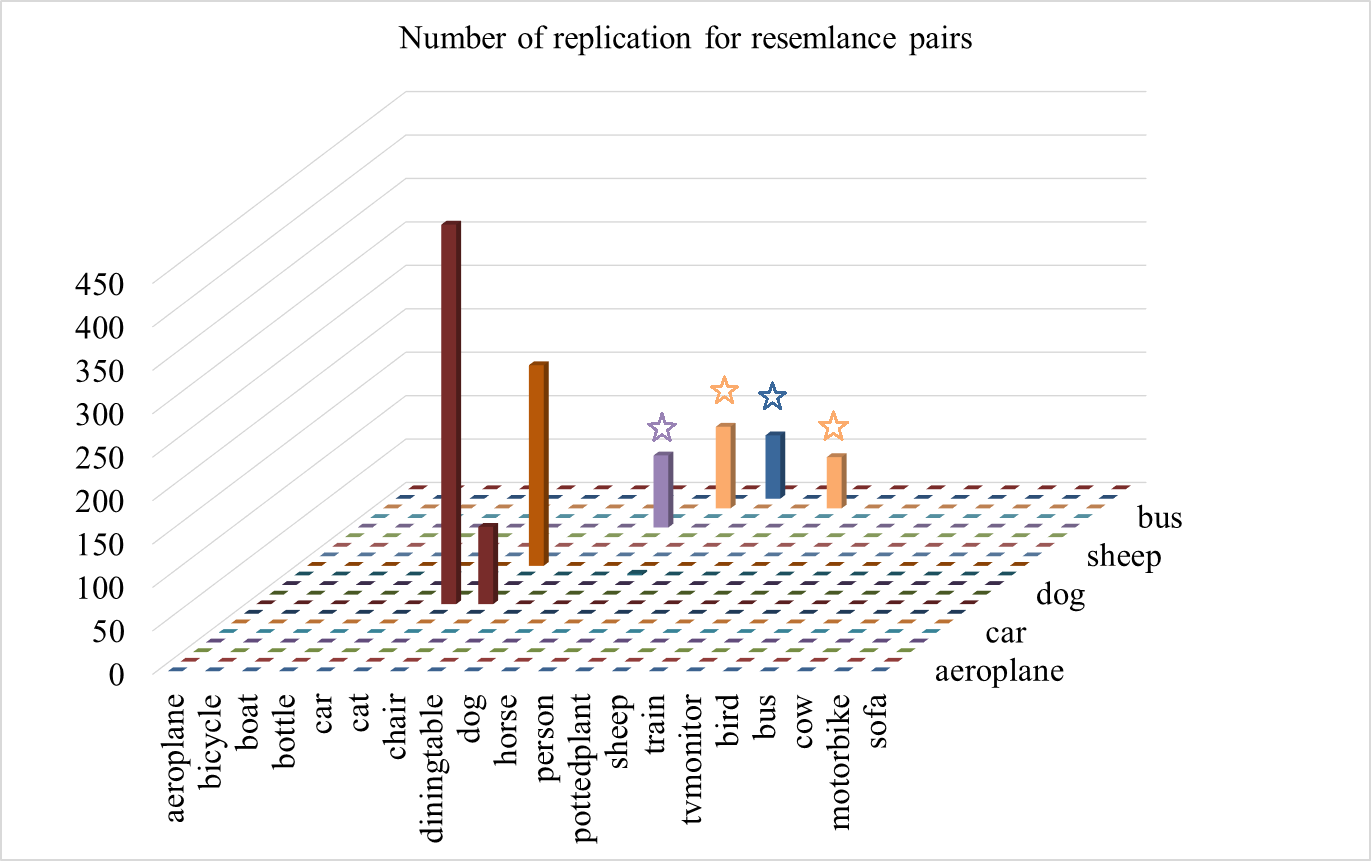}
	\caption{The number of replication of the resemblance pairs that constitute resemblance group. This result is under PASCAL VOC experiment at iteration milestone of 75\%. Notation \FiveStarOpen above the bars indicates the retained resemblance pairs which have at least one novel class.}
	\label{fig:Replication threshold}
\end{figure}

{\bf (\romannumeral2) Replication threshold.} To evaluate the confidence of the resemblance pair, we set a number for the replication threshold ($Th_{Rep}$). Once the number of how many times a resemblance pair appears ($ToR$) exceeds the replication threshold ($ToR\!>\!Th_{Rep}$) and contains novel classes in it during a certain number of training iterations, we record and save these pairs in a resemblance group: $G_{R} = \left\lbrace \left\langle C^{i}_{p}, C^{j}_{gt}\right\rangle, ...,\left\langle C^{k}_{p}, C^{m}_{gt}\right\rangle \right\rbrace \Rightarrow \left\lbrace C_{1}, C_{2}, ..., C_{n}\right\rbrace$, where $C_{1}, C_{2}, ..., C_{n}$ are unduplicated class labels. We visualize the number of replication of the $P_{R}$ and the corresponding $G_{R}$ in Fig.~\ref{fig:Replication threshold}. Some apparent replications occur between pairs $P_{R} = \left\langle bird, dog\right\rangle$, $P_{R} = \left\langle cow, sheep\right\rangle$, $P_{R} = \left\langle cow, horse\right\rangle $, and $P_{R} = \left\langle motorbike, bicycle\right\rangle$, in which case,  $G_{R}$ would be $\left\lbrace bird, dog, cow, sheep, horse, motorbike, bicycle \right\rangle$.

%-------------------------------------------------------------------------
{\bf Apply RCL to resemblance group.} During the process of a training batch, for a proposal from RoI head, if and only if either its predicted class $C^{i}_{p}$ or corresponding ground truth class $C^{j}_{gt}$ belongs to the resemblance group ($\left\lbrace C^{i}_{p}, C^{j}_{gt}\right\rbrace \cap G_{R} \neq \varnothing$), it can be passed through the contrastive encoder to get the corresponding contrastive feature $f_{Con}(p)$. Afterward, these features of proposals $f_{Con}(p)$ are compared with each other using cosine similarity.

%-------------------------------------------------------------------------
{\bf Iteration milestone.} The model should possess a certain capability of recognizing novel class objects. Applying RCL in the middle of the training process instead of at the very beginning or the end of it could help the model gets a basic capability of recognizing the novel objects after several rounds of iterations. We assume that the result is affected by the iteration milestone ($I_m$), which means the percentage of the whole training iterations has gone through. Therefore, we set and observe various iteration milestones to obtain the best iteration milestone. Once the current training iteration ($I_c$) reaches the iteration milestone, i.e. $I_c \geqslant I_m$, the RCL procedure is applied; before that, we use the general contrastive learning instead, as shown in Fig.~\ref{fig:flowchart-b}.

%-------------------------------------------------------------------------
\subsection{RCL loss}

We design the refined contrastive learning loss ($L_{RCL}$) focusing on $G_{R}$ for few-shot object detection.
\begin{equation}
L_{FSRC} = L_{cls} + L_{Bbox} + L_{objectness} + \lambda L_{RCL}
\label{eq:loss 1}
\end{equation}

Equation \eqref{eq:loss 1} shows the overall loss $L_{FSRC}$, which consists of classification loss $L_{cls}$, box regression loss $L_{Bbox}$, RPN loss $L_{objectness}$ and the RCL loss $L_{RCL}$ , also shown in Fig.~\ref{fig:flowchart-a}, where $\lambda$ is the balancing factor for the refined contrastive learning. Equation \eqref{eq:RCL loss 2} is the averaged weighted loss of general supervised contrastive learning loss $L_{z_i}$. $u_i$ is the $IoU$ of a proposal with its corresponding ground truth; $w(u_i) = I\{u_i \geq \phi \} \cdot g(u_i)$ is a cut-off function that calculates the weight applied on $L_{z_i}$ of each proposal pair, where $\phi=0.7$ is an $IOU$ threshold to filter out the proposals with low $u_i$. Re-weight function $g(u_i)$ is usually used as a hard-clip value and set to 1.

\begin{equation}
L_{RCL}=\frac{1}{N_{RCL}}\sum_{i=1}^{N_{RCL}}w(u_i)\cdot\!L_{z_i}
\label{eq:RCL loss 2}
\end{equation}
\begin{equation}
\begin{split}
N_{RCL}=\left\{
\begin{array}{ll}
\!N_{all},&{I_c < I_m}\\
\!N_{G_{R}},&{I_c \geqslant I_m}\\
\end{array} \right.
\label{eq:RCL loss 3}
\end{split}
\end{equation}
{\small
\begin{equation}
L_{z_i}\!=\!\frac{-1}{N_{y_i}\!-\!1}\!\sum_{j=1,j\neq i}^{N_{RCL}}\!I\{y_i= y_j\}\cdot\log\!\frac{e^{\tilde{z_i} \cdot\!\tilde{z_j}\!/\!\tau}}{\sum_{k=1}^{N_{RCL}}\!I_{k\neq i}\cdot e^{\tilde{z_i}\cdot\tilde{z_k}/\!\tau}}
\label{eq:RCL loss 4}
\end{equation}
}

$N_{RCL}$ in \eqref{eq:RCL loss 2} represents the total number of RoI proposals, where $N_{all}$ and $N_{G_{R}}$ are the number of all proposals and the number of proposals that belongs to $G_{R}$, respectively in \eqref{eq:RCL loss 3}; $I_c$ and $I_m$ represent the current iteration and the iteration milestone. In \eqref{eq:RCL loss 4}, $z_i$ depicts the normalized feature of the $i$-th proposal, which is used to measure the cosine similarity between proposals. $N{y_i}$ is the number of proposals with the same label as the $i$-th region proposal $y_i$. $I\{*\}$ in all equations is a binary truncation function that equals to 1 if the condition $*$ is satisfied, and equals to 0 if not.

\begin{table*}[ht]
\begin{center}
{\caption{Few-shot object detection performance on PASCAL VOC dataset. The novel classes mAP50 are evaluated on three separate splits. Our proposed FSRC reaches new SOTA in most of the scenarios. \textbf{The highest mAP50 for each column} are in black bold text, and \textcolor{blue}{\underline{the second highest scores}} is in blue text with underline.}\label{tab:voc}}
	\centering
	\resizebox{\textwidth}{!}{
		\begin{tabular}{l|l|lllll|lllll|lllll}
			\toprule
			\multirow{2}{*}{\diagbox[height=20pt,innerrightsep=25pt]{Method}{Shot}} & \multirow{2}{*}{Backbone} & \multicolumn{5}{c|}{Split1} & \multicolumn{5}{c|}{Split2} & \multicolumn{5}{c}{Split3} \\
			& & 1 & 2 & 3 & 5 & 10 & 1 & 2 & 3 & 5 & 10 & 1 & 2 & 3 & 5 & 10\\
			\midrule
			FRCN-ft \cite{Wang19} & \multirow{12}{*}{FRCN-R101} & 13.8 & 19.6 & 32.8 & 41.5 & 45.6 & 7.9 & 15.3 & 26.2 & 31.6 & 39.1 & 9.8 & 11.3 & 19.1 & 35.0 & 45.1 \\
			FRCN+FPN-ft \cite{Wang20} & & 8.2 & 20.3 & 29.0 & 40.1 & 45.5 & 13.4 & 20.6 & 28.6 & 32.4 & 38.8 & 19.6 & 20.8 & 28.7 & 42.2 & 42.1 \\
			TFA w/ fc \cite{Wang20} & & 36.8 & 29.1 & 43.6 & 55.7 & 57.0 & 18.2 & 29.0 & 33.4 & 35.5 & 39.0 & 27.7 & 33.6 & 42.5 & 48.7 & 50.2 \\
			TFA w/ cos \cite{Wang20} & & 39.8 & 36.1 & 44.7 & 55.7 & 56.0 & 23.5 & 26.9 & 34.1 & 35.1 & 39.1 & 30.8 & 34.8 & 42.8 & 49.5 & 49.8 \\
			MPSR \cite{Wu20} & & 41.7 & - & \textcolor{blue}{\underline{51.4}} & 55.2 & 61.8 & 24.4 & - & 39.2 & 39.9 & 47.8 & 35.6 & - & 42.3 & 48.0 & 49.7 \\
			FSCE \cite{Sun21} & & 44.2 & 43.8 & \textcolor{blue}{\underline{51.4}} & \textbf{61.9} & 63.4 & 27.3 & 29.5 & \textcolor{blue}{\underline{43.5}} & 44.2 & 50.2 & \textcolor{blue}{\underline{37.2}} & \textcolor{blue}{\underline{41.9}} & \textcolor{blue}{\underline{47.5}} & \textcolor{blue}{\underline{54.6}} & \textcolor{blue}{\underline{58.5}} \\
			Retentive R-CNN \cite{Fan21} & & 42.4 & 45.8 & 45.9 & 53.7 & 56.1 & 21.7 & 27.8 & 35.2 & 37.0 & 40.3 & 30.2 & 37.6 & 43.0 & 49.7 & 50.1 \\
			DC-Net \cite{Hu21a} & & 33.9 & 37.4 & 43.7 & 51.1 & 59.6 & 23.2 & 24.8 & 30.6 & 36.7 & 46.6 & 32.3 & 34.9 & 39.7 & 42.6 & 50.7 \\
			TIP \cite{Li21a} & & 27.7 & 36.5 & 43.3 & 50.2 & 59.6 & 22.7 & 30.1 & 33.8 & 40.9 & 46.9 & 21.7 & 30.6 & 38.1 & 44.5 & 50.9 \\
			CME \cite{Li21b} & & 41.5 & \textcolor{blue}{\underline{47.5}} & 50.4 & 58.2 & 60.9 & 27.2 & 30.2 & 41.4 & 42.5 & 46.8 & 34.3 & 39.6 & 45.1 & 48.3 & 51.5 \\
			FSOD-UP \cite{Wu21} & & 43.8 & \textbf{47.8} & 50.3 & 55.4 & 61.7 & \textcolor{blue}{\underline{31.2}} & \textcolor{blue}{\underline{30.5}} & 41.2 & 42.2 & 48.3 & 35.5 & 39.7 & 43.9 & 50.6 & 53.5 \\
			KFSOD \cite{Zhang22} & & \textcolor{blue}{\underline{44.6}} & - & \textbf{54.4} & 60.9 & \textbf{65.8} & \textbf{37.8} & - & 43.1 & \textbf{48.1} & \textcolor{blue}{\underline{50.4}} & 34.8 & - & 44.1 & 52.7 & 53.9 \\
			\midrule
			GCL (Our Impl.) & & 44.3 & 42.2 & 51.0 & 60.2 & 63.2 & 28.1 & 31.1 & 44.3 & 44.8 & 51.2 & 36.7 & 44.7 & 47.6 & 55.3 & 57.5 \\
			\rowcolor{mygray} \textbf{FSRC (Ours)} & \multirow{-2}{*}{FRCN-R101} & \textbf{45.5} & 43.4 & 51.1 & \textcolor{blue}{\underline{61.4}} & \textcolor{blue}{\underline{64.0}} & 28.4 & \textbf{31.3} & \textbf{45.0} & \textcolor{blue}{\underline{46.1}} & \textbf{51.6} & \textbf{38.8} & \textbf{45.1} & \textbf{48.4} & \textbf{55.5} & \textbf{59.0} \\
			\bottomrule
		\end{tabular}
  }
\end{center}
\end{table*}

\section{Experiments}
\label{sec:experiments}

%-------------------------------------------------------------------------
\subsection{Training details}

Faster R-CNN with a Resnet-101 feature pyramid network is the benchmark. 4 GPUs are used to train the model with a batch size of 16. We use stochastic gradient descent as the optimizer while setting the momentum to 0.9 and the weight decay to 0.0001. The initial learning rate is set as 0.001, we use the pre-trained model presented by \cite{Wang20}. For both PASCAL VOC and COCO datasets, the Resnet-101 backbone and the RoI pooling layer parameters are frozen during the fine-tuning process. The evaluation protocols are the same as those in \cite{Wang20}. We use the same basic hyperparameters as in \cite{Sun21}.

%-------------------------------------------------------------------------
{\bf Implementation on PASCAL VOC.} We follow the previous work in \cite{Wang20} and divide the PASCAL VOC dataset into two parts: 15 base classes and 5 novel classes. The novel classes are selected diversely based on three random splits as depicted in \cite{Wang20}, namely split1, split2, and split3. For the few-shot setting, we follow the K-shot (K = 1, 2, 3, 5, 10) sampling rule as mentioned in \cite{Wang20}. Training images are sampled from both PASCAL VOC 2007 and 2012, while evaluate on PASCAL 2007. Average precision (AP50) is evaluated separately as base class AP (bAP50) and novel class AP (nAP50). During fine-tuning, the experimental resemblance group $IoU$ threshold ($Th_{IoU}^{G_{R}}$) is set to 0.1 for K = 1,2 shots and set to 0.5 for K = 3, 4, 5 shots. The replication threshold is set to 0. The iteration milestone is set to 75\%.

%-------------------------------------------------------------------------
{\bf Implementation on COCO.} Following the previous works in \cite{Wang20, Sun21}, 60 categories in the COCO dataset are used as base classes, and other 20 categories are applied as novel classes. Images are trained with K-shot (K = 10, 30). We evaluate both AP and AP75 for COCO dataset. The resemblance group $IoU$ threshold ($Th_{IoU}^{G_{R}}$) is empirically set to 0.5 for both 10-shot and 30-shot. Replication threshold is set as 0 and iteration milestone is set to 50\%.

%-------------------------------------------------------------------------
\subsection{RCL results} 

{\bf Results on PASCAL VOC.} Experimental results on PASCAL VOC are listed in Tab.~\ref{tab:voc}, which shows the comparison of our proposed method with the most recent years' research work on the novel class AP (nAP50) over the three random splits. It obviously indicates our method outperforms the SOTA results on PASCAL VOC dataset in most cases, and especially surpasses all shots in split3. Meanwhile, in Tab.~\ref{tab:voc stddev} we list and compare the standard deviations between our FSRC method and general supervised contrastive learning (GCL) on both novel classes and all classes in different split, which demonstrate that our proposed RCL method lowers standard deviation among all classes and novel classes in the majority of the scenarios.

\begin{table}
\begin{center}
	{\caption{AP50 standard deviation on VOC of novel classes and all classes.}\label{tab:voc stddev}}
	\centering
	\begin{tabular}{l|l|l|l|l|l|l}
		\toprule
		& Shot & 1 & 2 & 3 & 5 & 10 \\
		\midrule
		\multirow{2}{*}{Split 1} & Novel & 13.3 & 8.4 & 10.5 & 11.5 & 12.2 \\
		& All & 17.6 & 17.9 & 15.3 & 12.6 & 13.3 \\
		\midrule
		\multirow{2}{*}{Split 2} & Novel & 12.3 & 15.1 & 13.9 & 15.5 & 14.0 \\
		& All & 24.5 & 23.7 & 17.1 & 17.8 & 16.1 \\
		\midrule
		\multirow{2}{*}{Split 3} & Novel & 17.9 & 17.2 & 18.6 & 18.0 & 17.7 \\
		& All & 21.6 & 19.4 & 18.1 & 15.8 & 15.4 \\
		\bottomrule
	\end{tabular}
\end{center}
\end{table}

\begin{table}
\begin{center}
	{\caption{Few-shot object detection performance on COCO dataset. Evaluation for novel classes AP and AP75 are listed. Our results have the highest scores above most previous works. \textbf{The highest AP for each column} is in black bold text, and \textcolor{blue}{\underline{the second highest scores}} are in blue text.}\label{tab:coco}}
	\centering
	\resizebox{0.4\textwidth}{!}{
		\begin{tabular}{l|ll|ll}
			\toprule
			\multirow{2}{*}{\diagbox[innerrightsep=15pt]{Method}{Shot}}&\multicolumn{2}{c|}{Novel AP}&\multicolumn{2}{c}{Novel AP75}\\
			&10&30&10&30\\
			\midrule
			TFA w/ cos \cite{Wang20}&10.0&13.7&9.3&13.4\\
			FSCE \cite{Sun21}&11.9&\textcolor{blue}{\underline{16.4}}&\textcolor{blue}{\underline{10.5}}&\textbf{16.2}\\
			SVD \cite{Wu21a}&\textcolor{blue}{\underline{12.0}}&16.0&10.4&15.3\\
			SRR-FSD \cite{Zhu21} &11.3&14.7&9.8&13.5\\
			N-PME \cite{Liu22}&10.6&14.1&9.4&13.6\\
			FORD+BL \cite{Nguyen22}&11.2&14.8&10.2&13.9\\
			\midrule
			GCL (Our Impl.)&11.0&16.1&9.5&15.0\\
			\rowcolor{mygray}\textbf{FSRC (Ours)}&\textbf{12.0}&\textbf{16.4}&\textbf{10.7}&\textcolor{blue}{\underline{15.7}}\\
			\bottomrule
		\end{tabular}
	}
\end{center}
\end{table}

%-------------------------------------------------------------------------
{\bf Results on COCO.} The results on COCO are listed in Tab.~\ref{tab:coco}. AP and AP75 of novel classes on both 10-shot and 30-shot are listed. Our results outperform most of the previous work and reach SOTA. Similarly, we compare the standard deviation of our results with the general supervised contrastive learning. Our results indicate that our proposed RCL significantly eliminates the polarization among all classes and novel classes, as shown in Tab.~\ref{tab:coco stddev}.

%-------------------------------------------------------------------------
\subsection{Ablation studies of hyperparameters}
\label{sec:ablation}

In this part, the experiments are performed on PASCAL VOC split1 5-shot scenario.

{\bf Iteration milestone.} The influence of iteration milestone ($I_m$) is studied by setting to different values in training stage to the overall training iterations: 5\%, 25\%, 50\%, 75\% and 100\%. $IoU$ threshold is set as 0.5, and the replication threshold is fixed as 5. The training results are listed in the second row of Tab.~\ref{tab:Ablation}, which indicates how different iteration milestones affect the training result inevitably. From the Tab.~\ref{tab:Ablation}, it is obvious that the setting of milestone to 75\% performs the best.

\begin{table}
\begin{center}
{\caption{AP standard deviation on COCO of novel classes and all classes.}\label{tab:coco stddev}}
	\centering
	\begin{tabular}{l|l|l}
		\toprule
		& 10 shot & 30 shot \\
		\midrule
		Novel classes & 9.2 & 10.2 \\
		All classes & 15.0 & 14.7 \\
		\bottomrule
	\end{tabular}
\end{center}
\end{table}

{\bf $\mathbf{IoU}$ threshold.} For the $IoU$ threshold ($Th_{IoU}^{G_{R}}$), we scale it from 0.0 to 0.9 to represent varies matching confidences of a proposal box with its ground truth box. The iteration milestone is fixed as 75\% and the replication threshold is fixed as 5. Naturally, higher ($Th_{IoU}^{G_{R}}$) will filter out more overlapped box pairs, and therefore reduce the quantity of resemblance pairs ($P_{R}$). Experimental results in the third row in Tab.~\ref{tab:Ablation} indicates that $Th_{IoU}^{G_{R}} = 0.5$ is a relatively moderate setting for this scenario.

{\bf Replication threshold.} The fourth row in Tab.~\ref{tab:Ablation} lists the results of different replication thresholds ($Th_{Rep}$) under $IoU$ threshold of 0.5. Higher $Th_{Rep}$ in charge of sorting out the frequently occurred resemblance pairs, and meanwhile ignore the pairs which has lower frequency of occurrence. Experimentally we use $Th_{Rep}$ as 0 to keep as many the resemblance pairs as possible.

\begin{table}
\begin{center}
{\caption{Ablation studies on hyperparameters.}\label{tab:Ablation}}
	\centering
	\begin{tabular}{l|l|l|l}
		\toprule
		$I_m$ & $Th_{IoU}^{G_{R}}$ & $Th_{Rep}$ & nAP \\
		\midrule
		5\% & \multirow{6}{*}{Fixed 0.5} & \multirow{6}{*}{Fixed 0} & 60.7 \\
		25\% & & & 60.7 \\
		50\% & & & 60.8 \\
		75\% & & & \textbf{61.4} \\
		100\% & & & 59.6 \\
		\midrule
		\multirow{5}{*}{Fixed 75\%} & 0.0 & \multirow{5}{*}{Fixed 0} & 61.1 \\
		& 0.25 & & 60.5 \\
		& 0.5 & & \textbf{61.4} \\
		& 0.75 & & 61.0 \\
		& 0.9 & & 60.0 \\
		\midrule
		\multirow{4}{*}{Fixed 75\%} & \multirow{4}{*}{Fixed 0.5} &0 & \textbf{61.4} \\
		& & 5 & 60.7 \\
		& & 10 & 61.0 \\
		& & 20 & 60.5 \\
		\bottomrule
	\end{tabular}
\end{center}
\end{table}

%------------------------------------------------------------------------
\section{Conclusion}
\label{sec:conclusion}

In this paper, we propose a novel end-to-end few-shot object detection model with a refined contrastive learning (FSRC) method. The introduced refined contrastive learning module can automatically find out the resemblance group and helps the network focusing on training this group at later stage of the fine-tuning process. Resemblance group is also proved to be existed in current benchmarks PASCAL VOC and COCO. The experimental results prove that FSRC outperforms current few-shot object detection in different shots and benchmarks. Furthermore, FSRC reduces standard deviation of average precision for both novel and base classes, which contributes to more stable prediction for different classes.

\bibliographystyle{IEEEtran}
\bibliography{main.bib}

\end{document}